\documentclass{MITcsail}

\usepackage{hyperref}
\definecolor{CiteGray}{HTML}{777777} % dark enough for contrast
\definecolor{CiteSlate}{HTML}{475569}  % subtle slate
\hypersetup{
  linkcolor=black,            % section refs, ToC
  citecolor=black,         % \citep, \citet
  urlcolor=black,          % \url, \href
  linktoc=page,                % only page numbers are linked in ToC
}

\usepackage{float,setspace}

\title{Invisible Languages of the LLM Universe}

\author
{Saurabh Khanna~$^{1, 2}$\footnote{Correspondence E-mail: s.khanna@uva.nl}, Xinxu Li~$^{1}$\\
\vspace{1em} % Space between authors and afilliations
\normalfont{\small $^{1}$Amsterdam School of Communication Research, University of Amsterdam}\\
\normalfont{\small $^{2}$Pembroke College, University of Oxford}\\
}

\begin{document}

\maketitle
\thispagestyle{firstpagestyle} % Draws the header on the first page

\begin{abstract}
Large Language Models are trained on massive multilingual corpora, yet this abundance masks a profound crisis: of the world's 7,613 living languages, approximately 2,000 languages with millions of speakers remain effectively invisible in digital ecosystems. We propose a critical framework connecting empirical measurements of language vitality (real-world demographic strength) and digitality (online presence) with postcolonial theory and epistemic injustice to explain why linguistic inequality in AI systems is not incidental but structural. Analyzing data across all documented human languages, we identify four categories: \textit{Strongholds} (33\%, high vitality and digitality), \textit{Digital Echoes} (6\%, high digitality despite declining vitality), \textit{Fading Voices} (36\%, low on both dimensions), and critically, \textit{Invisible Giants} (27\%, high vitality but near-zero digitality) -- languages spoken by millions yet absent from LLM training data. We demonstrate that these patterns reflect continuities from colonial-era linguistic hierarchies to contemporary AI development. Our analysis reveals that English dominance in AI is not a technical necessity but an artifact of power structures that systematically exclude marginalized linguistic knowledge. We conclude with implications for decolonizing language technology and democratizing access to AI benefits.
\end{abstract}

\onehalfspacing
\setlength{\parskip}{.5em}

\section{Introduction}

An examination of the availability of digital content in Javanese -- a language with more than 69 million speakers -- reveals minimal representation on platforms that are central to the training of Large Language Models (LLMs). By contrast, Icelandic, with approximately 320,000 speakers, exhibits a disproportionately large digital presence. This apparent paradox underscores a fundamental problem in the way contemporary artificial intelligence systems capture and model human linguistic diversity. Over the past five years, LLMs such as GPT-4, Claude, and Gemini have expanded rapidly, achieving performance that approximates human-level proficiency across a broad array of tasks \citep{achiam2023gpt, anthropic2024claude}. These models are trained on massive multilingual corpora harvested from the internet, including Common Crawl’s roughly 300 billion web pages, Wikipedia’s more than 64 million articles, and large-scale digital text repositories \citep{gao2020pile}. However, this ostensible abundance of data obscures a substantial structural bias: the overwhelming majority of training material originates from fewer than twenty high-resource languages, while most of the world’s 7,613 documented languages receive negligible coverage and are effectively rendered \textit{invisible} in these models \citep{eberhard2015ethnologue, joshi2020state}.

Prior scholarship has characterized this crisis from multiple perspectives. \citet{kornai2013digital} estimated that only approximately 5\% of the world’s languages are likely to attain substantive digital vitality, and introduced the concept of “digital language death” to denote language extinction in online environments as a problem distinct from, and additional to, conventional offline endangerment. \citet{bender2021dangers} showed that large language models encode and propagate hegemonic worldviews, with the attendant harms disproportionately affecting marginalized communities. Complementarily, \citet{simons2022assessing} proposed automated methodologies to quantify digital language support at a global scale, reporting that only 33 languages (0.4\%) exhibit full digital functionality across 143 surveyed platforms.

However, the predominant body of research conceptualizes linguistic exclusion primarily as a problem of technical resource scarcity, wherein languages are classified as `low-resource' on the basis of insufficient training data. This framing, however, obscures more fundamental lines of inquiry: \textit{why} do some languages with millions of speakers possess virtually no digital footprint? Which historical, political, and structural dynamics give rise to this inequality? Moreover, how does the divergence between on-the-ground prevalence (or \textit{vitality}) and online or computational prevalence (or \textit{digitality}) both reflect and reproduce colonial and neocolonial power relations?

We address these research questions through an integration of large-scale empirical measurement and critical theoretical frameworks. Drawing on a comprehensive dataset that quantifies linguistic vitality and digital presence for all 7,613 documented languages, we build postcolonial linguistic theories from \citet{said1977orientalism} and \citet{levisen2019postcolonial}, conceptualizations of epistemic injustice \citep{fricker2007epistemic, helm2024diversity}, and critical analyses of linguistic imperialism \citep{phillipson1992linguistic}. This combined approach enables us to investigate the underlying structural, ideological, and historical mechanisms that explain \textit{why} certain languages remain marginalized or rendered invisible in digital and epistemic spaces despite substantial demographic strength.

Our analysis yields three primary contributions. First, we demonstrate that the vitality-digitality gap is not neutral; rather, it systematically disadvantages languages spoken in formerly colonized regions. Invisible Giants -- languages exhibiting high sociolinguistic vitality but near-zero digital presence -- are disproportionately concentrated in Africa, South Asia, Southeast Asia, and Indigenous regions of the Americas, thereby recapitulating historical patterns of colonial domination. We conceptualize this as a form of \textit{digital epistemic injustice}, namely, the systematic exclusion of marginalized linguistic communities from AI-mediated processes of knowledge production. Second, we trace continuities from colonial-era linguistic practices to contemporary large language model (LLM) development. Missionary linguistics frequently construed languages as discrete, extractable objects; subsequent ISO coding regimes further reified them as fixed and enumerable entities; contemporary natural language processing (NLP) methodologies inherit and operationalize these epistemic assumptions. The prevailing `low-resource' characterization itself reproduces colonial logics, positioning African and Asian languages as inherently deficient rather than interrogating the political-economic conditions under which digital and computational infrastructures were never developed for them. Third, we articulate concrete implications for the decolonization of language technologies. In contrast to approaches that focus primarily on `adding more data' for underrepresented languages, we argue for a fundamental reconfiguration of LLM development practices. This includes the use of community-governed datasets, the design of alternative evaluation metrics that prioritize contextualized, non-majority linguistic features, and mechanisms for redistributing the economic value generated by AI systems to the language communities whose data underpins model training.

The situation presents a clear urgency: as large language models increasingly function as core infrastructure for education, commerce, and governance, the vitality of languages in digital environments is becoming a critical determinant of their long-term survival. If current development trajectories persist, artificial intelligence systems are more likely to intensify than mitigate processes of language endangerment, thereby establishing self-reinforcing feedback loops in which digital marginalization contributes directly to accelerated offline linguistic decline.

\section{Theoretical Framework}

\subsection{Digital Language Death and the Vitality-Digitality Distinction}

\citet{lewis2010assessing} introduced the EGIDS (Expanded Graded Intergenerational Disruption Scale) as a systematic framework for assessing language vitality and endangerment, extending the original GIDS (Graded Intergenerational Disruption Scale) proposed by \citet{fishman1991reversing}. Building on this line of inquiry, \citet{kornai2013digital} demonstrated that digital vitality constitutes an autonomous dimension, largely decoupled from traditional measures of language vitality. Thus, a language may exhibit robust intergenerational transmission and therefore rank highly on EGIDS, while simultaneously being effectively absent from the digital sphere (i.e., exhibiting negligible or zero digital presence). Drawing on a large-scale dataset comprising 8,426 languages, Kornai proposed a four-tier typology of digital status and showed that languages classified as \textit{heritage} -- those that are digitally documented but no longer employed in active, everyday communication -- have a very low probability of achieving higher levels of digital presence, irrespective of the extent of documentation efforts.

This distinction exposes a fundamental limitation in existing language endangerment assessment frameworks. The EGIDS evaluates offline language vitality primarily on the basis of intergenerational transmission and domains of institutional use, assigning languages to categories ranging from 0 (International) to 10 (Extinct). Similarly, UNESCO’s six-level framework emphasizes dimensions such as speaker attitudes, language policies, and the extent of linguistic documentation \citep{unesco2003language}. However, both frameworks were envisioned prior to the point at which the internet became a central medium for economic activity and social participation.

Recent scholarship has sought to address this discrepancy. The Digital Language Diversity Project proposed a six-level language vitality scale that evaluates capacity, presence, and performance across multiple digital domains \citep{soria2018digital}. In parallel, \citet{simons2022assessing} introduced an automated evaluation framework encompassing seven support categories (from content presence to modern speech assistance) and 143 technological platforms. Nonetheless, no existing framework has yet succeeded in systematically integrating grassroots and digital dimensions of language presence and strength into a coherent model that has been empirically validated across a broad range of linguistic contexts.

We employ a two-dimensional analytical framework in which vitality and digitality are modeled as orthogonal axes. \textit{Vitality} captures the strength of a language in grassroots terms, operationalized via the estimated number of first-language speakers (sourced from Ethnologue) and its current EGIDS classification. \textit{Digitality} quantifies the language’s presence in digital environments, measured through its representation in Common Crawl (159 billion web pages analyzed), Wikipedia (64 million articles), Hugging Face datasets (114{,}000) and models (447{,}000), as well as open language archives (474{,}000 entries). 
For both vitality and digitality, we aggregate the respective indicators using factor analysis to obtain single standardized vitality and digitality scores. On this basis, we define a \textit{representation score} as the difference between the digitality and vitality scores, which serves as an approximation of the discrepancy between a language’s sociolinguistic strength and its digital representation.

\subsection{Epistemic Injustice in Language Technology}

The notion of language modeling bias introduced by \citet{helm2024diversity} denotes built-in preferences for particular languages within AI systems. They argue that this bias constitutes a form of epistemic injustice: such systems tend to exhibit high precision for languages associated with dominant social groups, while being substantially less capable of representing socio-culturally salient concepts specific to marginalized communities. Large language models (LLMs) thus have the potential to reproduce and amplify both testimonial injustice (the unjust deflation of a speaker’s credibility due to identity prejudice) and hermeneutical injustice (structural gaps in shared interpretive resources that hinder the understanding of certain social experiences) at an unprecedented scale. Testimonial injustice manifests when models systematically downrank, mistranslate, or fail to recognize contributions articulated in non-dominant languages. For example, an input in Swahili receives systematically lower-quality processing than an equivalent input in English, such that the system effectively treats the Swahili speaker's testimony as less reliable or less worthy of engagement. Hermenutical injustice occurs when the training data for a given language lack the lexical or conceptual resources necessary to articulate community-specific or domain-specific concepts. For instance, if isiZulu training data do not contain terms corresponding to dinosaurs or evolution, the LLM's ability to support, extend, or even enable scientific discourse in isiZulu is significantly constrained \citep{nekoto2020participatory}.

Recent scholarship extends this framework by arguing that designating particular data as `knowledge' within knowledge-enhanced large language models (LLMs) can obscure harms experienced by marginalized groups. In the absence of meaningful diversification, such knowledge-enhancement practices risk reproducing and amplifying epistemic injustice. For example, immigration and asylum adjudication processes that rely on narrowly scoped machine translation systems may place lives at risk when asylum seekers are unable to articulate their claims in languages that these systems recognize or adequately process. The opacity, scale, and infrastructural embeddedness of contemporary AI systems enable epistemic injustice on an unprecedented magnitude: billions of interactions per day can implicitly communicate that certain languages, epistemic traditions, and modes of knowing are inconsequential or invalid.

We extend this line of inquiry by situating it within the framework of digital language vitality. The discrepancy between a language's sociolinguistic vitality and its level of digital representation constitutes a direct source of hermeneutical injustice. Specifically, communities with robust oral traditions and large speaker populations are systematically excluded from the benefits of artificial intelligence because their linguistic resources have not been digitized or incorporated into computational infrastructures. This condition does not reflect incidental scarcity, but rather a form of structural exclusion rooted in, and perpetuating, historical patterns of marginalization.

\subsection{Postcolonial Linguistics and Digital Imperialism}

\citet{said1977orientalism} conceptualized Orientalism as a discursive formation through which Western societies exert dominance over, and systematically reconfigure, Eastern societies, representing them as underdeveloped and inferior. Building on this, \citet{ismail2024revisiting} demonstrate how linguistic hierarchies established during colonial rule continue to shape the development of modern technologies. The predominance of English in AI systems is thus not a neutral or purely technical outcome, but rather a manifestation of historical imperial expansion and the ensuing technological hegemonies that structure contemporary knowledge production. 

Contemporary language technologies exhibit extractive dynamics in which institutions located predominantly in the Global North appropriate data from communities in the Global South, develop computational models based on these data, and subsequently commercialize and resell the resulting tools to those same communities for profit \citep{mohamed2020decolonial}. Phenomena such as the production of algorithmic bias in facial recognition systems \citep{ruha2019race}, data annotation performed under exploitative labor conditions \citep{gray2019ghost}, and community-led struggles for data sovereignty \citep{kukutai2016indigenous} illustrate multiple lines of continuity between historical forms of territorial colonialism and contemporary data colonialism. These processes reproduce classical patterns of colonial resource extraction and reinforce longstanding power asymmetries between the West and the Global South in the domain of language, now intensified and reconfigured through the deployment of artificial intelligence in the digital age.

The very designation of a `low-resource language' can be understood as reproducing a colonial epistemic framework. As \citet{adebara2022towards} argue, many African languages are currently classified as low resource not due to any inherent deficiency, but because Western institutions have historically failed to invest in the digital infrastructure necessary for their computational development. This terminology thereby constructs these languages as intrinsically lacking and in need of benevolent support from already well-resourced institutions, rather than foregrounding the fact that resource scarcity is a politically and historically produced outcome of systemic marginalization.

\citet{migge2025material} examine historical-material continuities between early missionary linguistics and contemporary natural language processing. Missionary-era construction of `languages' as discrete, extractable, and ostensibly bounded objects predates modern NLP’s conceptualization of languages as datasets to be collected, segmented, and exploited. ISO-639-3 language codes, developed within Western standardization regimes, similarly presuppose languages as stable, clearly delimited entities rather than as dynamic, heterogeneous, and inherently fluid sociolinguistic practices. This ontological framing facilitates large-scale computational processing while simultaneously obscuring and erasing the complexity of lived linguistic repertoires. Moreover, contemporary commercial language technologies continue to rely heavily on colonial-era translations and grammatical descriptions produced by missionaries, thereby reproducing and extending the epistemic and material effects of those earlier regimes of linguistic knowledge production.

Self-orientalization further intensifies these dynamics. \citet{yan2018other} document how certain Chinese professionals strategically mobilize and even amplify cultural stereotypes, self identifying with Confucian traits, in order to secure advantages in professional settings. Similarly, \citet{chen2021language} demonstrate that tourism discourse in China not only constructs particular regions as peripheral on the basis of limited English-language services, but also reiterates tropes of Chinese English deficiency. This discursive pattern constitutes a form of internalized linguistic imperialism that is likely to be replicated and amplified by large language models (LLMs) trained on such corpora.

\subsection{Language Ideology and Technological Design}

Language ideologies -- belief systems concerning language that serve to legitimize and naturalize existing social structures -- are necessarily instantiated within technological infrastructures \citep{woolard2020language}. Contemporary conversational AI systems operationalize Western linguistic ideologies that privilege specific modes of language use, including standardized grammar over vernacular varieties, written over oral forms, explicitness over contextual inference, and monolingualism over multilingual repertoires. These ideological commitments inform core design decisions regarding what is classified as high-quality language data, which linguistic features are prioritized for optimization, and which criteria are employed to assess system performance.

The ideology of a `standard language' constructs socially prestigious varieties as normatively correct while marking other varieties as deficient or erroneous \citep{lippi2012english}. Large language models (LLMs) are trained predominantly on formal written registers, thereby internalizing and reproducing prescriptive linguistic norms. When such models `correct' African American Vernacular English into Standard American English, they operationalize and reinforce raciolinguistic ideologies that frame the former as improper or broken \citep{blodgett2020language}. Analogous processes occur in multilingual contexts: what is labeled as `correct' Hindi is typically formal written Hindi, which marginalizes and effectively erases Hinglish and diverse regional varieties that constitute the everyday linguistic practices of millions of speakers.

Platform architectures embed and reproduce particular linguistic assumptions. For example, Twitter’s character constraints structurally privilege logographic writing systems relative to alphabetic ones. TikTok's emphasis on audio‑visual content tends to favor languages that already possess extensive, globally circulating popular media repertoires. LinkedIn's normative expectations of a professional register advantage languages with long-established conventions of formal business communication. Despite their widespread influence, these platform-specific language ideologies have received limited scholarly attention; systematic analyses are still lacking regarding how platform affordances differentially advantage or disadvantage specific languages and communicative practices.

\section{Methods}

\subsection{Data Sources and Coverage}

We analyze a comprehensive dataset encompassing all 7,613 languages documented in the 25th edition of Ethnologue \citep{eberhard2015ethnologue}. The analysis integrates two primary dimensions.

\emph{Vitality} is operationalized by combining:

\begin{itemize}
    \item First-language speaker counts from Ethnologue, which range from single-digit values (for languages on the verge of extinction) to several hundred million speakers (e.g., Hindi, Arabic, Spanish)
    \item EGIDS (Expanded Graded Intergenerational Disruption Scale) status ratings, spanning from 0 (International) to 10 (Extinct), which evaluate the degree of intergenerational transmission and the extent of institutional support and development
\end{itemize}

\emph{Digitality} is quantified by aggregating a language’s presence across multiple large-scale digital resources:

\begin{itemize}
    \item Common Crawl: 159 billion web pages (November 2023 release)
    \item Wikipedia: 64 million articles distributed across more than 300 language editions
    \item Hugging Face: 114{,}000 datasets and 447{,}000 models annotated with language tags
    \item Open Language Archives Community (OLAC): 474{,}000 catalogued language resource entries
\end{itemize}

Following \citet{simons2022assessing}, we implement automated language identification using the fastText and CLD3 classifiers, complemented by validation sampling. For each language, we calculate normalized prevalence scores across each of the aforementioned corpora and subsequently perform principal component analysis–based dimensionality reduction to derive composite indices of linguistic vitality and digital presence.

The degree of visibility of a language on digital platforms is quantified by its \textit{representation score}, which is defined as:
\begin{equation}
\text{Representation} = \text{Digitality}_{\text{normalized}} - \text{Vitality}_{\text{normalized}} \, .
\end{equation}
Negative values of this score identify languages whose sociolinguistic vitality exceeds their level of digital presence, indicating underrepresentation in online environments. Conversely, positive values correspond to languages that are overrepresented in the digital sphere relative to their speaker base.

\noindent
We classify languages into four categories based on their position in vitality-digitality space:

\begin{itemize}
    \item \textbf{Strongholds} (+ Vitality / + Digitality): Entities characterized by both vitality and digitality scores exceeding the sample median.
    \item \textbf{Digital Echoes} (-- Vitality / + Digitality): Entities exhibiting below-median vitality combined with above-median digitality.
    \item \textbf{Fading Voices} (-- Vitality / -- Digitality): Entities for which both vitality and digitality scores fall below the median.
    \item \textbf{Invisible Giants} (+ Vitality / -- Digitality): Entities displaying above-median vitality in conjunction with below-median digitality.
\end{itemize}

\section{Results}

\subsection{The Four-Category Framework}

\begin{figure}[H]
    \centering
    \includegraphics[width=0.9\textwidth]{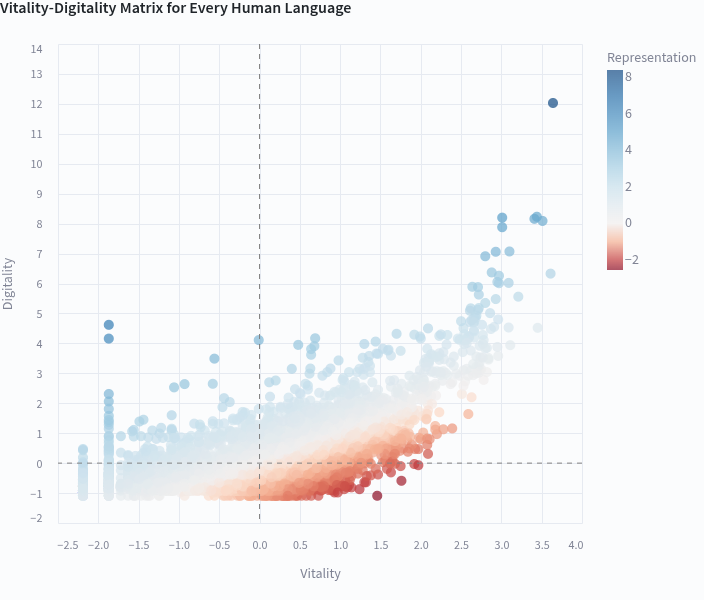}
    \caption{Mapping 7613 human languages on vitality (ground presence) and digitality (web presence).}
    \label{fig:overall}
\end{figure}

\begin{figure}[H]
    \centering
    \includegraphics[angle=90,width=.8\textwidth,height=\textheight]{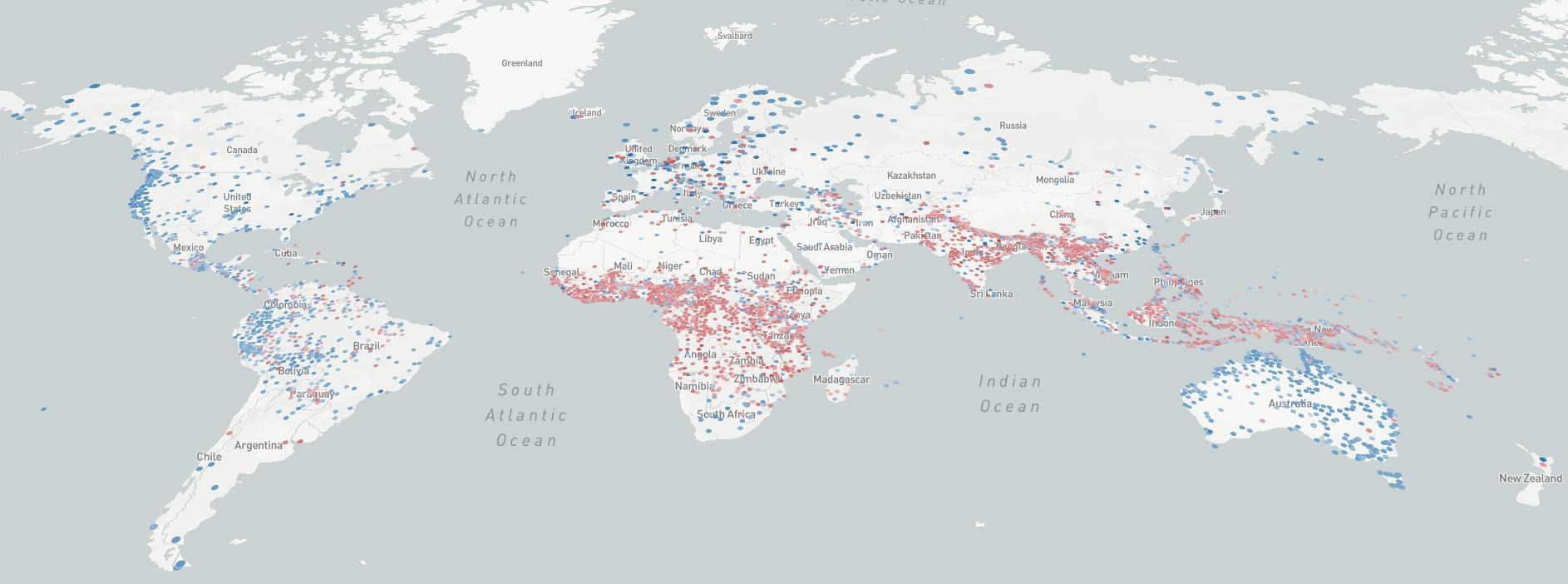}
    \caption{Geolocating invisible languages}
    \label{fig:map}
\end{figure}

Application of the information visibility framework to all 7,613 documented human languages reveals a pronounced systemic disparity: patterns of linguistic representation within digital ecosystems diverge substantially from patterns of demographic vitality. Figure \ref{fig:overall} depicts the resulting vitality–digitality matrix, in which each language is positioned according to its real-world demographic strength (x-axis) and its aggregated digital presence across web pages, Wikipedia, datasets, computational models, and archival resources (y-axis). The representation score -- defined as the difference between digitality and vitality -- quantifies this discrepancy, with blue values indicating digital over-representation and red values indicating systematic under-representation.
Four distinct clusters emerge, each revealing different dynamics of visibility and erasure. The geographic spread of these languages can be seen \ref{fig:map}, highlighting the stark colonial legacies of invisibility in parts of Africa, Asia, and Oceania.

\clearpage

\clearpage
\begin{figure}[H]
    \centering
    \includegraphics[width=0.9\textwidth]{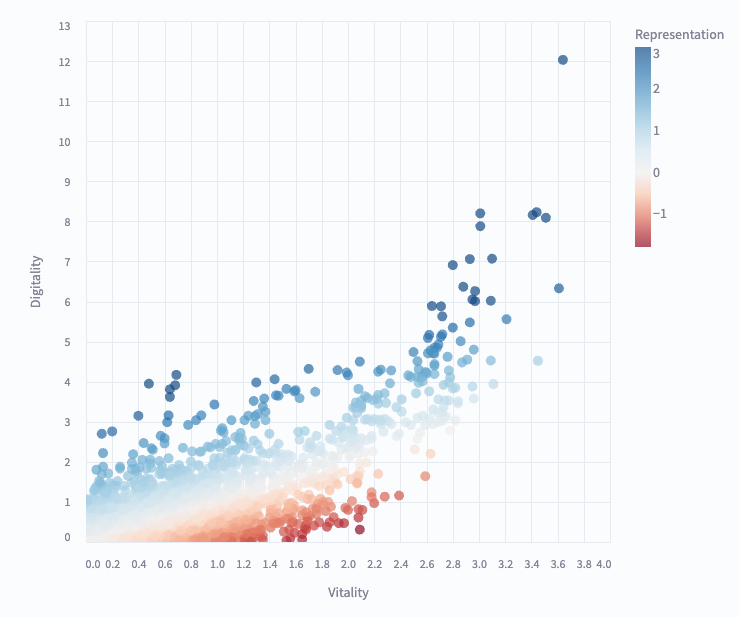}
    \caption{Strongholds}
    \label{fig:strongholds}
\end{figure}

\textbf{Strongholds (+ Vitality / + Digitality)}: Approximately one-third of the languages in the sample fall within the upper-right quadrant, characterized by both high demographic vitality and substantial digital representation (see Figure \ref{fig:strongholds}). These languages appear as a plume extending above both zero-reference axes, with speaker populations, institutional backing, and online presence functioning as mutually reinforcing factors. This group comprises not only global lingua francas but also regionally dominant languages that have effectively converted offline sociolinguistic strength into robust digital visibility. From the standpoint of large language models (LLMs), these languages already constitute the bulk of training corpora and serve as primary reference points for multilingual evaluation benchmarks. The principal policy implication is thus one of sustained maintenance rather than active remediation, ensuring that available data remain current while reallocating capacity-building efforts and resources to languages situated in more vulnerable quadrants.

\clearpage
\begin{figure}[H]
    \centering
    \includegraphics[width=0.9\textwidth]{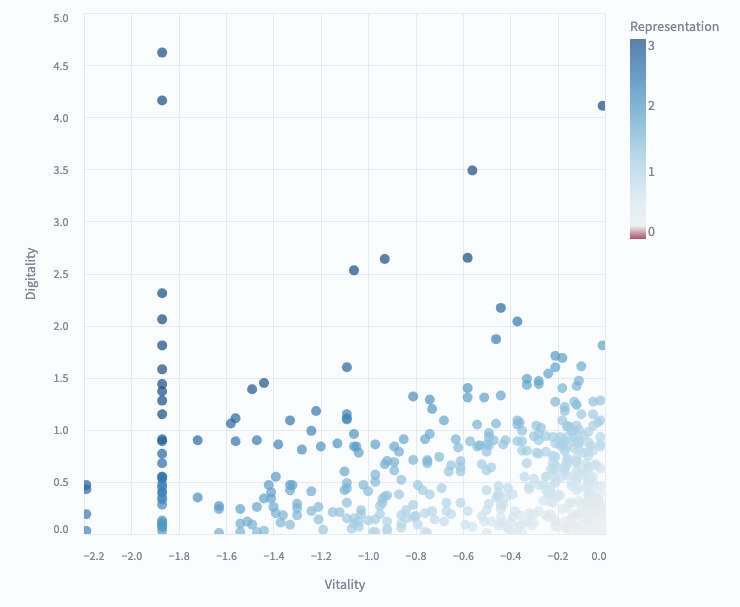}
    \caption{Digital Echoes}
    \label{fig:digechoes}
\end{figure}

\textbf{Digital Echoes (-- Vitality / + Digitality)}: Only 6\% of languages are located in the upper-left quadrant, representing cases in which digital presence exceeds that of shrinking speaker communities (see Figure \ref{fig:digechoes}). Typically associated with historical prestige, liturgical or ritual functions, and/or active diaspora networks, these languages illustrate that digitally available corpora can persist beyond the decline of face-to-face communicative practices. In the visualization, they appear as a thin blue band above the vitality origin, emphasizing how documentation, standardization, and archiving can preserve linguistic material even as inter-generational transmission and everyday use diminish.

From a research perspective, Digital Echoes underscore a critical methodological limitation: elevated digital indicators (e.g., corpus size, number of online documents, or web traffic) do not equate to robust linguistic health. This disconnect arises because most widely used digital metrics are usage-agnostic with respect to (i) the age structure and size of the speaker population, (ii) the domains of actual use (home, school, government, workplace, etc.), (iii) transmission to younger generations, and (iv) speakers’ language attitudes, identity functions, and community control over language resources. A language may thus exhibit substantial digital output generated by a small, aging, or highly specialized group (such as clergy, scholars, or heritage enthusiasts), or by nonnative learners and institutions, while the number of fluent, everyday speakers continues to contract. Conversely, automated or large-scale digitization (e.g., of historical texts) can inflate digital volume without indicating current communicative vitality or community-driven use. For policymakers, these languages constitute strategic leverage points: existing digital infrastructures -- including corpora, standardized orthographies, pedagogical materials, and online communities -- can be mobilized in support of revitalization policies, such as the development of curricular resources, digital learning tools, and community media, prior to further erosion of any remaining active fluency and functional domains of use.

\clearpage
\begin{figure}[H]
    \centering
    \includegraphics[width=0.9\textwidth]{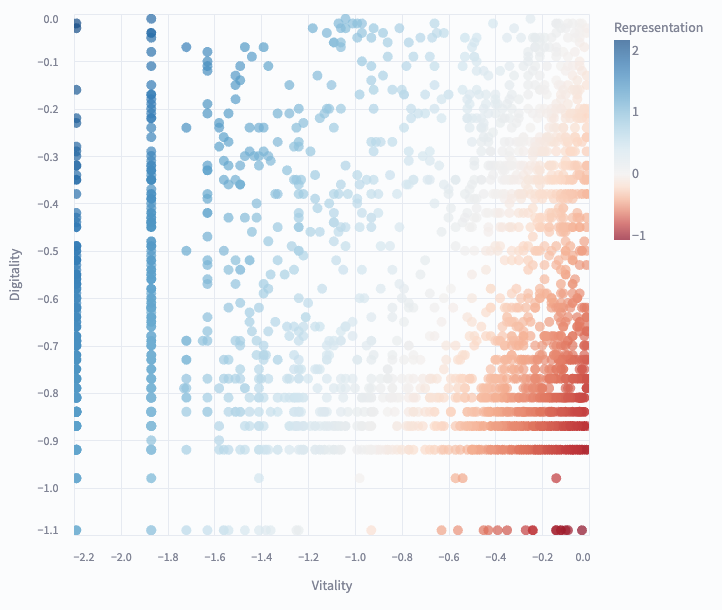}
    \caption{Fading Voices}
    \label{fig:fadingvoices}
\end{figure}

\textbf{Fading Voices (-- Vitality / -- Digitality)}: The densest cluster, comprising approximately 2,700 languages (36\%), is concentrated near the origin of the space and extends into the lower-left quadrant (see Figure \ref{fig:fadingvoices}). These languages exhibit a dual deficit: they are characterized by small, vulnerable speech communities and extremely limited digital presence. In the visualization, they appear as a diffuse cloud of nearly colorless points, indicating that when both dimensions are severely constrained, even the representation-gap metric yields minimal informative signal. From the perspective of large language models (LLMs), these languages are effectively absent from the training landscape. From the perspective of linguistic diversity and language endangerment, however, they represent the bleeding edge of potential loss: without rapid, systematic field documentation and community-driven efforts to record and maintain linguistic practices, these languages are at high risk of disappearing beyond the possibility of empirical reconstruction.

\clearpage
\begin{figure}[H]
    \centering
    \includegraphics[width=0.9\textwidth]{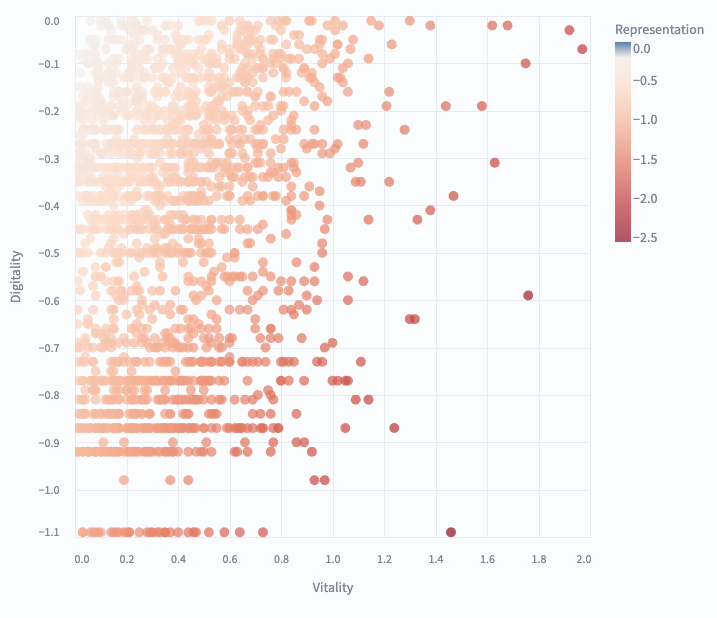}
    \caption{Invisible Giants}
    \label{fig:invisgiant}
\end{figure}

\textbf{Invisible Giants (+ Vitality / -- Digitality)}: Approximately 2,000 languages (27\%) are located in the lower-right quadrant, which constitutes the primary area of analytical concern in this study (see Figure \ref{fig:invisgiant}). These languages appear as intensely red points: their vitality extends toward the positive end of the horizontal axis, while their digitality remains below the horizontal reference line, yielding the largest positive "need for action" differential on the color scale. These are languages with millions of active speakers but minimal or negligible digital representation. The paradox is striking: substantial demographic presence coexists with an almost complete absence from digital domains.

The Invisible Giants category encapsulates the central argument of this work: digital exclusion cannot be reduced to linguistic endangerment or to small speaker populations. Languages may be socio-linguistically robust, widely used, demographically significant, and transmitted across generations, while remaining systematically marginalized or absent within the digital ecosystems that increasingly structure authoritative knowledge production and shape the operation of contemporary AI systems.

\subsection{Theoretical Interpretation: Why Invisible Giants Exist}

The existence of Invisible Giants -- languages with millions of speakers yet minimal digital presence -- cannot be explained by technical resource constraints alone. If digitality simply reflected speaker numbers, the vitality-digitality plot would approximate a diagonal line. Instead, we observe systematic deviation: certain languages punch far above their demographic weight digitally, while others with larger populations remain nearly absent. This pattern aligns precisely with postcolonial theory's predictions. \citet{said1977orientalism} established how colonial powers constructed hierarchies positioning non-Western languages as inferior, requiring civilizing intervention. These hierarchies did not disappear with formal decolonization -- they became embedded in digital infrastructure decisions.

The infrastructure for digitality -- Unicode encoding, keyboard layouts, spell-checkers, search algorithms, content moderation systems -- was built primarily by and for Western languages. When platforms assess `notability' for Wikipedia articles, `quality' for search rankings, or `representativeness' for training data, they apply standards developed in English-speaking contexts. For search engines, these ostensibly neutral technical systems encode cultural assumptions that systematically disadvantage marginalized communities.
The concept of epistemic injustice sheds light on the mechanism. Digital systems commit hermeneutical injustice by failing to provide linguistic resources needed for marginalized communities to interpret and share their experiences. When an Invisible Giant language lacks digital presence, it is hard for the speakers access AI tools in their language, find online educational resources or just participate in digital commerce on equal footing.  The system renders their knowledge production illegible, positioning them as passive consumers of content produced in dominant languages rather than active knowledge creators.

Critically, the Invisible Giants category demonstrates that this exclusion is structural, not natural. These languages are not "low-resource" due to inherent deficiency -- they are systematically under-resourced by institutions that chose to invest in some languages but not others. The framing matters: `low-resource' naturalizes scarcity, while `systematically under-resourced' centers the political decisions that produced it \citep{adebara2022towards}.

\subsection{Implications for Large Language Models}

The vitality-digitality gap has profound implications for LLMs, which train on precisely the digital corpora our framework measures. When Invisible Giant languages constitute 27\% of languages with substantial speaker bases yet receive minimal representation in Common Crawl, Wikipedia, and scholarly archives, LLM training data systematically excludes nearly a third of the world's demographically robust linguistic diversity.

This exclusion produces three compounding harms:

\begin{itemize}
    \item \textbf{Performance inequality}: Models trained predominantly on Stronghold languages perform dramatically worse on Invisible Giants. This performance gap is not merely inconvenient, but also determines who can access AI-powered education, commerce, healthcare, and governance. As LLMs become infrastructure, linguistic exclusion becomes infrastructural inequality.
    \item \textbf{Knowledge erasure}: LLMs encode not just linguistic patterns but conceptual structures, cultural knowledge, and ways of knowing embedded in training text. When training data excludes Invisible Giants, models cannot represent the knowledge systems, historical narratives, and lived experiences of communities speaking those languages. This constitutes epistemic violence at scale -- billions of parameters encoding the message that certain languages and knowledge don't matter.
    \item \textbf{Feedback loops}: Poor LLM performance discourages speakers from using their languages digitally, reducing digital content generation, which further reduces training data, which worsens performance. This creates self-reinforcing cycles where invisibility begets further invisibility -- a digitally mediated language shift mechanism operating at unprecedented scale and speed.
\end{itemize}

The geographic map in Figure \ref{fig:map} provides a spatial representation of these patterns. Much of Africa appears in red, indicating languages whose sociocultural vitality exceeds their degree of digital representation. Many Sub-Saharan African languages, despite being sustained for centuries through robust oral traditions and complex knowledge systems, remain structurally underrepresented or entirely absent in digital repositories. By contrast, much of Europe appears in blue, reflecting that even relatively small European languages exhibit a level of digital presence that surpasses what would be expected from their speaker populations alone, largely due to long-term institutional investment in digital infrastructures. 
This distribution does not constitute a random pattern but rather reflects the re-articulation of colonial-era hierarchies through contemporary algorithmic systems. Analogous to how colonial regimes extracted material resources while representing colonized populations as lacking civilization, current AI development practices frequently rely on training data derived from digital labor concentrated in the Global South, while simultaneously classifying non-Western languages as `low-resource' and framing them as objects of technocratic or philanthropic intervention.

\subsection{Validation Through Existing Research}

Our empirical results both corroborate and refine prior, largely fragmentary, observations. \citet{kornai2013digital} estimated that only 5\% of the world’s languages are likely to attain substantive digital vitality and argued that digital extinction proceeds largely independently of offline endangerment trajectories. The four-category framework we propose supplies the theoretical architecture that Kornai’s analysis implicitly called for. In particular, Digital Echoes operationalize his claim that languages preserved in archival resources may exhibit a form of digital presence without corresponding contemporary communicative use, while Invisible Giants instantiate his finding that absence from Wikipedia is a strong predictor of a language’s inability to attain higher levels of digital presence.

\citet{simons2022assessing} conducted a large-scale assessment of digital language support across 143 platforms and found that only 33 languages (0.4\%) attained full functional capabilities. While their methodology primarily emphasizes the dimension of technical infrastructure support, our framework introduces an additional vitality perspective, demonstrating that the mere availability of infrastructure does not ensure actual language use. Within our typology, some Stronghold languages exhibit both robust infrastructure and active user communities; Invisible Giants may possess comparable infrastructural provision yet display minimal content production; and Fading Voices are characterized by the absence of both adequate infrastructure and sustained usage.

The distributional pattern characterized as the 27\% Invisible Giants empirically substantiates the concerns articulated by \citet{bender2021dangers}, namely that the paucity of LLM training data for the majority of the world’s languages is not accidental but structurally embedded in current data collection and curation pipelines. The `stochastic parrots' critique -- according to which LLMs reproduce and amplify biases at scale without genuine understanding -- acquires heightened significance when we quantify which specific 27\% of demographically robust languages are systematically marginalized or omitted from training corpora.

Crucially, our framework sheds light on an aspect that prior work has not quantified: the magnitude of the representation gap. It is not only important to document that certain languages lack adequate digital resources, but also to demonstrate that approximately 2000 languages, collectively spoken by millions of individuals, fall into the highest `need for action' category. This quantitative characterization reframes the issue from an anecdotal concern into a measurable crisis that necessitates systematic and large-scale intervention.

\section{Discussion}

\subsection{Digital-Epistemic Injustice as Structural Phenomenon}

Our findings demonstrate that the vitality-digitality gap is not a technical accident but a structural phenomenon reflecting power relations. The systematic concentration of Invisible Giants -- 27\% of all languages, representing roughly 2,000 languages with millions of speakers yet minimal digital presence -- reveals that digital exclusion operates independently of demographic strength. This constitutes a kind of \textit{digital epistemic injustice}: the systematic exclusion of marginalized linguistic communities from AI-mediated knowledge production through denial of both digital infrastructure and epistemic authority. This dual exclusion operates at unprecedented scale in LLM development. When billions of parameters encode patterns from training data that systematically excludes these 2000 languages, the resulting models embed the message that certain ways of knowing, certain knowledge systems, and certain communities don't matter.

The mechanism operates through self-reinforcing feedback loops. Initial digital inequality leads to training data scarcity, producing poor LLM performance, discouraging digital language use and reinforcing data scarcity. Meanwhile, Stronghold languages -- already enjoying institutional support and digital infrastructure -- attract platform investment, NLP research funding, and commercial applications, widening gaps. Without intervention, these dynamics accelerate language shift as younger generations conclude that economic participation requires adopting dominant languages.

Critically, this is not inevitable technological progress but the reproduction of colonial hierarchies through algorithmic means. The concentration of Strongholds in Europe (where even minority languages maintain robust digital presence) while Invisible Giants concentrate in formerly colonized regions reveals historical continuities. Colonial language policies positioned indigenous languages as inferior, requiring civilizing intervention; contemporary `low-resource language' framing positions them as deficient, requiring benevolent development assistance. Both framings naturalize inequality while obscuring its political origins.

\subsection{Limitations and Future Directions}

\paragraph{Methodological limitations} First, our digitality measurement relies on publicly accessible web data, potentially missing private communication in messaging applications. If Invisible Giant speakers use their languages extensively via WhatsApp, SMS, or encrypted platforms, our analysis underestimates actual digital use. However, from an LLM training perspective, private communication does not contribute to model development, so our measurement captures the dimension relevant for AI representation. Second, the four-category classification uses median splits, creating discrete boundaries where gradients exist. Languages near category boundaries may be misclassified, and the framework does not capture within-category variation -- not all Strongholds enjoy equal digital presence, nor do all Invisible Giants face identical challenges. Future work should develop continuous measures of the representation gap alongside categorical classifications. Third, we lack longitudinal data on how vitality-digitality gaps evolve. Are gaps widening or narrowing? Do digital interventions successfully narrow gaps for targeted languages? Cross-sectional analysis cannot answer these causal questions. Longitudinal tracking of language trajectories across the vitality-digitality space would enable causal inference about what drives languages from one quadrant to another.

\paragraph{Theoretical extensions} Future research should examine intersections of linguistic marginalization with other forms of oppression. How does digital language inequality interact with special needs (especially for signed languages' digital representation)? With gender, particularly for languages where speaker populations are gendered? With age, as youth language innovations occur primarily in digital spaces? These intersectional analyses remain underdeveloped.
Platform-specific language ideologies warrant deeper investigation. Different platforms encode different linguistic norms -- Twitter's character limits advantage logographic systems, TikTok's audio-visual primacy benefits languages with media presence, LinkedIn's professional register expectations advantage languages with established business communication. How do platform affordances systematically advantage or disadvantage specific languages? We need granular socio-technical analysis of how architectural choices embed linguistic assumptions.
Community-centered research methodologies remain rare. Our study, like most, analyzes patterns from external positions without deep engagement with affected language communities in research design or interpretation. Developing genuine co-production methodologies where speakers shape research questions, interpret findings, and control data governance represents a methodological frontier essential for decolonizing research practice.

\paragraph{Empirical extensions} The framework's generalizability across domains requires testing, since we only developed it for languages, while similar frameworks can extend towards assessing low visibility of certain armed conflicts and scholarly research. Does the vitality-digitality logic extend to other knowledge domains -- traditional medicine, land management practices, oral historical archives? Systematic testing across domains would validate or refine theoretical claims about digital filtering mechanisms.
Regional deep dives would complement global analysis. Why do specific Invisible Giants remain underrepresented despite demographic strength? Detailed case studies examining language-specific barriers -- orthographic debates, institutional politics, diaspora dynamics -- would illuminate mechanisms obscured in aggregate analysis. Combining computational measurement with ethnographic depth would strengthen explanatory power here.

\subsection{Practical Implications}

\paragraph{For AI Developers} Current LLM training treats language representation as a data availability problem -- scrape more web pages, digitize more books. Our analysis reveals this approach is insufficient. The 2,000 Invisible Giants have speakers, they have knowledge to contribute, but they lack digital infrastructure and institutional support that would enable content creation at scales sufficient for LLM training. Targeted investment in digital infrastructure for Invisible Giants is required: orthography standardization where needed, keyboard interfaces, spell-checkers, text-to-speech systems. These are prerequisites for content creation, not merely nice-to-have additions. Community controlled data trusts enabling speakers to govern how language data is collected, used, and monetized would shift power dynamics from extractive to collaborative. Evaluation metrics must center non-English linguistic features -- tone marking accuracy, morphological complexity handling, pragmatic particle usage -- rather than English-centric benchmarks like BLEU scores that penalize linguistic differences as errors. Performance equity should be measured not just by aggregate metrics but by whether models serve speakers of Invisible Giants as effectively as they serve English speakers.

\paragraph{For Policymakers} Language planning must integrate digital dimensions. Granting a language official status without digital infrastructure investment produces hollow recognition. Policies should mandate that government digital services support languages spoken by significant populations, fund localization of open-source software and educational platforms, and establish accountability mechanisms requiring AI companies to report representation gaps and remediation efforts.
Digital Language Rights should extend existing linguistic human rights frameworks into digital domains. Just as the Universal Declaration of Linguistic Rights (1996) establishes rights to use one's language in education and public life, contemporary frameworks must recognize rights to digital language resources, AI services in one's language, and participation in knowledge production digitized by platforms.
International funding mechanisms should prioritize Invisible Giants for digital investment. These offer highest social return on investment -- millions of speakers flourish who could immediately benefit from digital resources.

\paragraph{For Researchers} Methodological decolonization requires fundamental shifts in how we conduct research. Co-designing studies with speaker communities rather than extracting data, compensating language consultants at professional rates, sharing model weights and tools with communities providing training data, publishing in open-access venues with translations into studied languages -- these practices should become norms, not exceptions.
Evaluating research impact by community-defined outcomes (language vitality metrics, economic benefits, cultural continuity) rather than only academic citations would better align incentives with social benefit. Research excellence should incorporate community endorsement as a criterion, recognizing that scholarship about marginalized communities should serve those communities' self-determined goals.
Funding agencies could accelerate these shifts by requiring visibility impact assessments for digitization grants -- explicitly addressing how proposed work will or will not narrow representation gaps -- and prioritizing proposals co-designed with affected language communities over top-down ones.

\subsection{Toward Linguistic Justice in AI}

The Invisible Giants paradox, populations numbering in the millions, whose speakers are effectively unheard in digital spaces -- encapsulates the core of the current crisis. Addressing this problem requires acknowledging a set of uncomfortable empirical and historical facts: the predominance of English in contemporary AI systems is neither an inherent linguistic inevitability nor a technical necessity. Rather, it is a manifestation of enduring geopolitical and socioeconomic power asymmetries that continue to determine which languages receive sustained investment, institutional recognition, and robust computational support.

From an optimistic standpoint, targeted and adequately resourced interventions could substantially reduce these disparities. The requisite methodological toolkit largely already exists: systematically quantify linguistic representation gaps; prioritize structurally marginalized, high-vitality languages; invest in community-governed digital and computational infrastructures; and design multilingual models with explicit guarantees of performance parity or fairness across languages.
From a realistic view though, prevailing incentive structures are misaligned with these goals. Commercial AI development is primarily optimized for high-income markets whose users predominantly speak globally dominant stronghold languages. Academic research agendas frequently prioritize incremental improvements on English-centric benchmarks, while multilingual and low-resource language work is often formulated as secondary or peripheral. Platform-based business models privilege engagement metrics that systematically favor content in already dominant languages. In the absence of deliberate structural interventions -- such as regulatory mandates, conditional public funding mechanisms, and sustained community-based organizing -- market dynamics are likely to further intensify linguistic and digital inequalities.

The resulting policy and design choice is stark: Will emerging AI technologies be deployed to democratize access to information, education, and economic opportunity across linguistic boundaries, or will they function as accelerants of language shift, compressing linguistic diversity into an increasingly English-centered monoculture? The outcome hinges on whether the gap between linguistic vitality and digital representation is conceptualized merely as a technical resource allocation problem, or as a form of digital and epistemic injustice. The latter framing underscores that closing this gap requires not only additional data and infrastructure, but also a more profound redistribution of power, resources, and epistemic recognition across linguistic communities.

\section{Conclusion}

We demonstrate that linguistic inequality in Large Language Models (LLMs) is not incidental but structurally embedded, reproducing historical continuities from colonial language hierarchies through to contemporary AI development. Of the world’s 7,613 attested languages, approximately 2,000 can be classified as Invisible Giants -- languages with millions of speakers that remain substantially underrepresented in digital ecosystems. This underrepresentation is disproportionately concentrated in formerly colonized regions, where rates of linguistic exclusion are estimated to be orders of magnitude higher than in Europe.

We conceptualize this configuration as a form of digital-epistemic injustice: the systematic marginalization of specific linguistic communities within AI-mediated regimes of knowledge production, enacted through constrained access to digital infrastructure and the systematic devaluation of their epistemic contributions. As LLMs increasingly function as core socio-technical infrastructure for education, commerce, governance, and cultural production, such exclusion is likely to intensify ongoing processes of language shift. This dynamic promotes convergence toward a homogenized, English-dominant model of multilingualism and, in turn, accelerates the erosion of global linguistic diversity.

This trajectory, however, is not immutable. Targeted and sustained investment in digital infrastructure for structurally under-resourced languages, the institutionalization of community-governed data stewardship, and the adoption of explicitly decolonial principles in AI design and governance have the potential to significantly mitigate these disparities within a generational timescale. The central question is political rather than purely technical: whether societies will accept AI systems that reproduce and normalize colonial linguistic hierarchies, or instead demand technological infrastructures that recognize, sustain, and amplify the full spectrum of human linguistic diversity.

The \emph{invisible languages} within the current LLM ecosystem need not remain invisible. Rendering them visible requires acknowledging that their exclusion has been historically and structurally produced rather than accidental, and recognizing that meaningful inclusion entails more than the quantitative expansion of training datasets. It necessitates a fundamental reconfiguration of whose interests AI systems are designed to serve and whose knowledge systems are accorded legitimacy and authority.

\small

\bibliographystyle{apalike}
\bibliography{references}

\section*{Acknowledgments}
We acknowledge research funding for this project from the Amsterdam School of Communication Research at the University of Amsterdam.

% \clearpage
% \beginsupplement

% \section{Supplementary Materials}
% Space to provide all supplementary materials used in our analysis.

\end{document}